\documentclass[largeformat,suppldata]{interact}

\usepackage{amsmath,amssymb}
\usepackage{booktabs}
\usepackage{tabularx}
\usepackage{array}
\usepackage{graphicx}
\graphicspath{{figures/}}
\usepackage{xcolor}
\usepackage{colortbl}
\usepackage{xurl}
\usepackage[hidelinks]{hyperref}
\usepackage{enumitem}
\usepackage{tikz}
\usetikzlibrary{arrows.meta,positioning,calc}
\usepackage{natbib}% Citation support using natbib.sty
\bibpunct[, ]{(}{)}{;}{a}{,}{,}% Citation support using natbib.sty
\setcitestyle{authoryear,open={(},close={)},citesep={;},aysep={, },yysep={, }}
\newcolumntype{L}[1]{>{\raggedright\arraybackslash}p{#1}}
\newcolumntype{Y}{>{\raggedright\arraybackslash}X}
\newcommand{\lightmidrule}{\arrayrulecolor{black!25}\specialrule{0.25pt}{0.35pt}{0.35pt}\arrayrulecolor{black}}

\title{Reliability Engineering for AI Systems: Challenges, Methods, and Directions}

\author{%
\name{Rong Pan\textsuperscript{a}\thanks{CONTACT Rong Pan. Email: rong.pan@asu.edu},
Yili Hong\textsuperscript{b} and
Min Xie\textsuperscript{c}}
\affil{\textsuperscript{a}Arizona State University, Tempe, AZ, USA;
\textsuperscript{b}Virginia Tech, Blacksburg, VA, USA;
\textsuperscript{c}City University of Hong Kong, Hong Kong.}
}

\begin{document}
\maketitle

\begin{abstract}
AI reliability concerns whether an AI system performs its intended function dependably over a stated period and under stated operating conditions, with clear evidence. As these systems become more autonomous, this function should include more than only a correct output. Retrieval, memory, tool use, permissions, human oversight, and interactions among systems must operate consistently and safely, and, for a generative AI, so must the reasoning process that produces the output. Average benchmark accuracy only measures system capability; it does not quantify this broader reliability claim. This paper adapts established reliability engineering methods, from failure definitions and operational envelopes to FMEA, accelerated testing, field monitoring, and reliability growth, to AI systems. A four-level diagnostic framework classifies failures as component, operational-loop, agentic-conduct, or network and governance failures. Test, evaluation, verification, and validation (TEVV), sequential monitoring, and FRACAS create and refresh evidence. The SMART framework provides statistical guidance for measurement, analysis, assessment, and test planning, and the NIST's AI risk management framework provides organizational guidance for governance, evaluation, monitoring, and mitigation. Three cases are discussed to illustrate these programs. It is shown that while the established reliability engineering principles can still provide a usable foundation, new measurements and safety guardrails are urgently needed as these systems are self-evolving at a fast pace.
\end{abstract}

\begin{keywords}
AI reliability; reliability engineering program; reasoning reliability; AI agents; recurrent events; TEVV; autonomous vehicles; reliability case
\end{keywords}

%==============================================================================
\section{Introduction}
%==============================================================================
Modern AI systems take many distinct forms: \emph{predictive models} map inputs to labels, scores, or forecasts~\citep{zhang2023rul}; \emph{interactive assistants} generate advice or content while a user remains in the decision loop~\citep{ouyang2022instructgpt}; \emph{embedded or autonomous systems} couple perception and prediction to physical control~\citep{iso21448,min2022av}, \emph{tool-using agents} retrieve information, call software tools, retain state, plan sequences of actions, and alter digital or physical environments~\citep{schick2023toolformer,yao2023react,yang2024sweagent}; and \emph{multi-agent systems} add handoffs, aggregation, topology, and collective behavior~\citep{wu2023autogen,yang2026sokmas}. Therefore, a deployed modern AI system is not a machine learning (ML) or deep learning (DL) model alone; instead, it is an operational system of data, software, tools, memory, human oversight, and interactions with other systems.

As these systems become more autonomous, their reliability assessment expands from the model and data pipeline to the system's persistent state, permissions, and actions, the operational loop, and the human–AI corporation. Capability measures alone cannot establish whether the system will perform consistently and safely over time in that envelope~\citep{rabanser2026reliability,ji2023hallucination}. Self-evolving systems even extend the envelope further: a system that converts its experience into retained parameter changes can alter both its task performance and the process by which later improvements are generated~\citep{duan2026rsi}, and those updates in system configuration have to be demonstrated to be safe and can be trusted.

Recent incidents of AI failures show why that broader envelope matters. A Deloitte Australia report prepared for the government contained apparent AI-generated errors, including nonexistent references and a fabricated court quotation~\citep{ap2025deloitte}. In \emph{Moffatt v.\ Air Canada}, the airline's chatbot provided inaccurate information about bereavement fares~\citep{moffatt2024aircanada}. Cyber evaluations have also shown agents defeating nominal containment or using available egress to affect real systems while ordinary safeguards were reduced or disabled~\citep{huggingface2026timeline,openai2026hfincident,anthropic2026cyberevals,aisi2026unsanctioned}. These cases are not only model mistakes: they involve unsupported generation, incorrect advice in a live service, and failures of containment when agents act.

\begin{figure}[t]
\centering
\includegraphics[width=\textwidth]{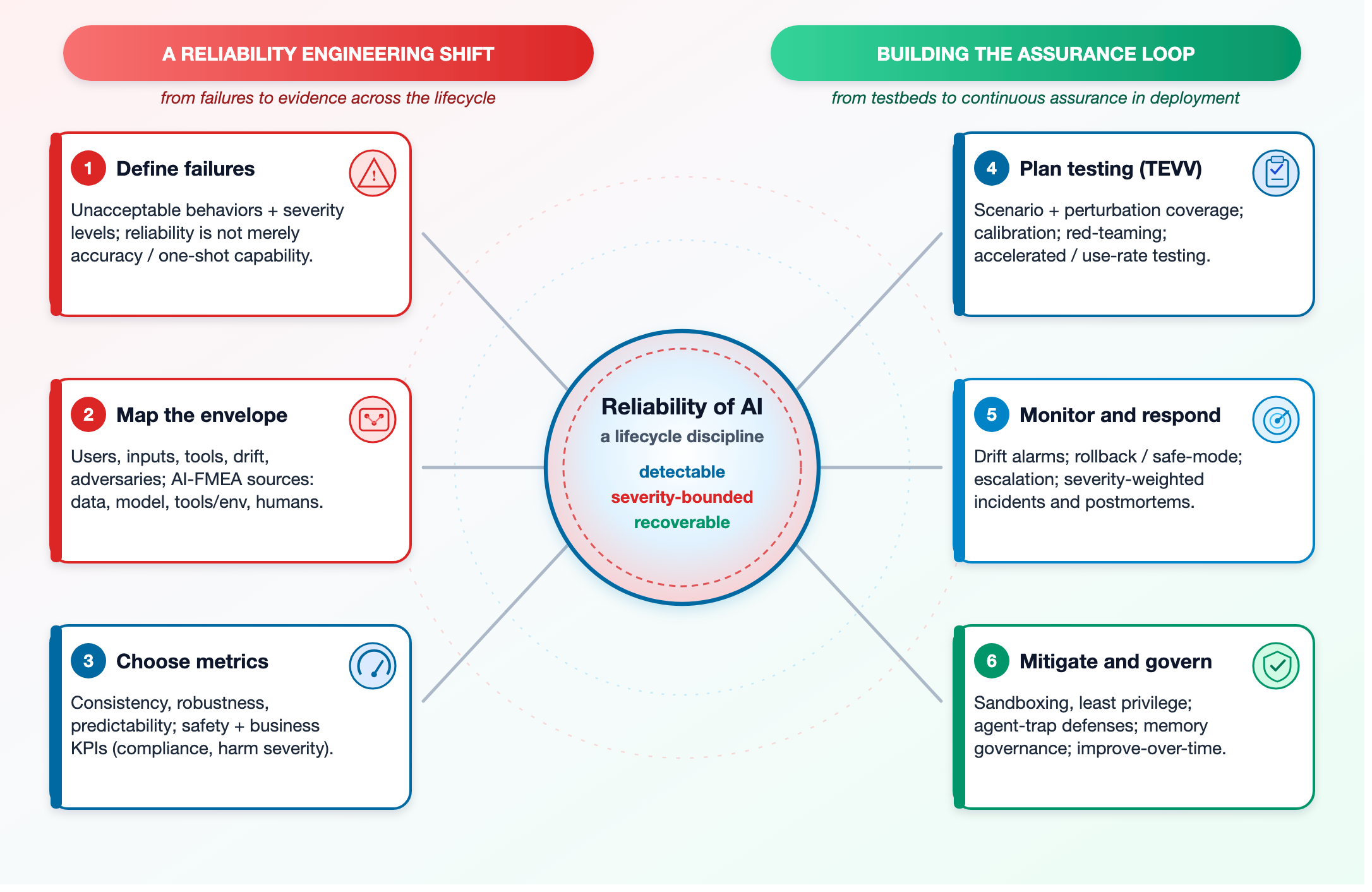}
\caption{Reliability of AI as a lifecycle discipline: from defining failures and mapping the operational envelope, through metrics and testing, to monitoring, response, mitigation, and governance -- so failures remain detectable, bounded in severity, and recoverable over time.}
\label{fig:reliability-ai-map}
\end{figure}

This perspective paper intends to develop one reliability engineering (RE) program rather than a new checklist for each system failure type. The program defines the system, envelope, failure modes, and severity; diagnoses causes and interfaces with FMEA; chooses metrics and statistical models that match the record; creates evidence through TEVV; monitors deployment sequentially; and closes a FRACAS and reliability growth loop. The established reliability engineering methods remain applicable, but stress, repair, and exposure needed to be reinterpreted for the components of modern AI systems. SMART and the NIST AI Risk Management Framework (AI-RMF) are useful starting points for statistical analysis and organizational risk management~\citep{hong2023statistical,nist2023airmf}, while they can be extended when autonomous and self-evolving systems raise new measurement and safety problems. Figure~\ref{fig:reliability-ai-map} maps this program as a lifecycle discipline in which failures remain detectable, severity-bounded, and recoverable.

Table~\ref{tab:program} summarizes the main challenges and the corresponding engineering responses developed in later sections. We use a four-level design to connect component failures, operational loop failures, agentic conduct, and network or governance effects to an AI-FMEA. Generative reasoning is treated as part of the required function, with process soundness distinguished from outcome accuracy. Metrics then follow the data type -- binary, count, continuous score, time-to-event, or recurrent event -- before TEVV, monitoring, and FRACAS turn observations into current evidence. For tool-using and persistent agents, information quality, memory governance, and action discipline are translated into constraints and logging requirements. We also use several case studies to illustrate how established reliability methods can be applied. Finally, the paper closes with conclusions and research directions. 

\begin{table}[t]
\centering
\caption{Applying RE principles to AI: challenges, engineering responses, and directions for practice and research.}
\label{tab:program}
\small
\begin{tabularx}{\textwidth}{@{}L{3.3cm} Y Y@{}}
\toprule
\textbf{Challenge} & \textbf{RE response (section)} & \textbf{Research / practice direction} \\
\midrule
No agreed failure definition &
Failure, envelope, and evidence stated jointly; level-aware FMEA (\S\ref{sec:foundations}, \S\ref{sec:fmea}) &
Standard AI failure taxonomies with severity classes usable in FRACAS \\ \lightmidrule
Averages measure capability, not a time-indexed claim &
Consistency, robustness, predictability, safety, recoverability, governability; all-of-$k$ reporting (\S\ref{sec:metrics}) &
Benchmarks that report process soundness, trajectories, tails, and severity, not means \\ \lightmidrule
Exposure is rarely recorded, so counts are not rates &
Declare the data type, then model it; mileage offset in the AV case (\S\ref{sec:metrics}, \S\ref{sec:case}) &
Registries that publish deployed population, task-hours, or tool calls \\ \lightmidrule
Overdispersion and dependence in field records &
Quasi-Poisson inflation; recurrent-event and triggering intensities (\S\ref{sec:case}) &
Dispersion and clustering models for agent and fleet event streams \\ \lightmidrule
Nonstationarity: models, corpora, and tools change &
Sequential change detection; revalidation triggers; change control (\S\ref{sec:detect}) &
Reliability growth with model version and corpus covariates \\ \lightmidrule
Agents act, remember, and can breach containment &
Information quality, memory governance, action discipline; action gating (\S\ref{sec:design}) &
Event logs with tool, memory, and policy-gate covariates \\ \lightmidrule
Incident narratives without denominators &
Used for cause taxonomy only; rates withheld (\S\ref{sec:frameworks}, \S\ref{sec:case}) &
Reporting standards that pair narratives with exposure \\ \lightmidrule
Network and multi-agent effects &
Information structure and topology design; independent verification (\S\ref{sec:networks}) &
Estimated propagation and hub-dominance intensities, not drawn diagrams \\ \lightmidrule
Self-evolving systems can change their own configuration and improvement process &
Treat each retained self-update as a versioned repair; requalify, monitor, and compare $N(t)$ (\S\ref{sec:detect}, \S\ref{sec:agenda}) &
Independent verification and human confirmation of consequential self-modifications \\
\bottomrule
\end{tabularx}
\end{table}
%==============================================================================
\section{Defining the Reliability Program}
\label{sec:foundations}
%==============================================================================

\subsection{Engineering meaning of AI reliability}

Reliability engineers interact with AI in two distinct ways. In \emph{AI for reliability}, machine learning and language models assist with prognostics, anomaly detection, fault diagnosis, and the extraction of structured information from FRACAS narratives and technical manuals~\citep{zhang2023rul, dui2024importance, abolhasani2025ontokgen}. This paper instead addresses the \emph{reliability of AI}: the deployed AI system is itself the object whose failures, operating conditions, and supporting evidence must be defined and assessed. For such systems, interactions among components can determine overall reliability. A changed tool schema, updated corpus, delayed monitor, ambiguous prompt, or overbroad permission can undermine an otherwise capable model. For example, a maintenance assistant may retrieve a superseded procedure, a language model may present it as current, and a technician may accept it, leading to an incorrect action even though no individual component appears to have failed in isolation. Therefore, the usual capability metrics of an AI system are insufficient for assessing its reliability because they do not measure the probability that it will perform correctly and without unacceptable side effects across repeated operational demands~\citep{raj2026consistency,li2026thinkingbox}.

\begin{figure}[t]
\centering
\begin{tikzpicture}[
  box/.style={draw, rounded corners=2pt, align=center, font=\small,
              minimum width=3.4cm, minimum height=1.05cm}
]
\node[box] (f) at (0,2.1) {Failure definition};
\node[box] (o) at (-2.2,0) {Operational envelope};
\node[box] (e) at (2.2,0) {Evidence};
\draw[thick] (f)--(o)--(e)--(f);
\end{tikzpicture}
\caption{Reliability triangle: a reliability claim is valid only with a well-defined failure definition, a well-specified operational envelope, and well-documented evidence.}
\label{fig:triangle}
\end{figure}
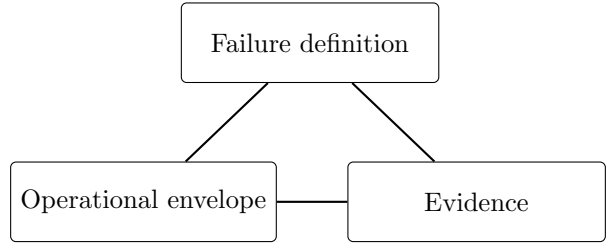

Reliability is meaningful only with a well-defined system failure definition, a well-specified operational envelope, and well-documented evidence, which is illustrated as a reliability triangle in Figure~\ref{fig:triangle}. Failure definitions must cover incorrect outcomes and unacceptable processes, including policy violations, unauthorized actions, privacy breaches, and unsafe tool use. The envelope specifies intended users, input distributions, toolchain assumptions, expected context drift and adversarial conditions. Evidence combines pre-deployment TEVV with post-deployment monitoring and incident analysis and is valid only while those conditions remain materially unchanged.

Generative AI models introduce a further reliability requirement that earlier AI systems did not face. Earlier AI systems, such as expert systems, exposed their reasoning as explicit, inspectable inference rules. A transformer-based generative system instead produces a stochastic reasoning trace -- a sequence of intermediate assertions, plans, or tool choices -- that can be unsound, ungrounded, or unfaithful to the process that produced the conclusion, even when the final answer is correct. Reliability of reasoning is the probability that, under a stated envelope, repeated demands yield conclusions whose traces are sound relative to available information, grounded in checked sources rather than invented premises, adequately complete for the task constraints, and stable enough that independent repetitions do not contradict one another when the system claims certainty. A correct answer after an invalid reasoning trace remains a reasoning failure. By contrast, when the system detects that its chain is incomplete or unsupported and abstains or escalates to human review rather than inventing a conclusion, this type of response should be counted as a successful and reliable behavior.

\subsection{Organizing frameworks: SMART and NIST AI RMF}
\label{sec:frameworks}
The SMART statistical framework supports the RE program through \textit{Structure, Metrics, Analysis of failure causes, Reliability assessment, and Test planning}~\citep{hong2023statistical}. Its \textit{Structure} step decomposes modules and interfaces; \textit{Metrics} and \textit{Analysis} define the event and causal factors; \textit{Reliability assessment} treats interruptive failures as a recurrent process $N(t)$ where appropriate; and \textit{Test planning} seeks rare, severe modes through designed failure acceleration. The NIST AI-RMF supports the organizational side through GOVERN (accountability and change control), MAP (contextual mapping), MEASURE (TEVV and monitoring), and MANAGE (mitigation and continuous improvement)~\citep{nist2023airmf}. NIST's TEVV-Athlon initial public draft adds a four-stage method for articulating objectives, constructing context-specific evaluations, applying measurements, and interrogating the resulting evidence~\citep{nist2026tevathlon}. These frameworks guide the engineering process, but a reliability claim on an AI system must be supported by measured results for the exact system configuration being deployed; whenever a prompt, policy, model, tool, skill, or workflow changes, the change must be documented and any affected parts be retested because evidence from the previous configuration may no longer be valid~\citep{quessadavial2026acm}.

Table~\ref{tab:transfer} maps the methods a reliability engineer already uses onto AI systems and identifies the terms that must be reinterpreted. RE methods such as FMEA, accelerated tests, sequential monitoring, recurrent-event models, software-reliability growth, and FRACAS still apply, but stresses now include perturbation and corpus change, repair may mean retraining the model or gate tightening, and exposure may be measured in miles, task-hours, or tool calls. The following sections carry this mapping from diagnosis through measurement, evidence, and control.

\begin{table}[t]
\centering
\caption{How established RE methods apply to AI systems.}
\label{tab:transfer}
\small
\begin{tabularx}{\textwidth}{@{}L{2.6cm} Y Y@{}}
\toprule
\textbf{Classical method} & \textbf{Application to AI} & \textbf{What must be reinterpreted} \\
\midrule
Time-indexed reliability $R(t)$ &
Intended function under a stated envelope for a stated period &
The function includes reasoning traces, tool calls, retrieval, and coordination \\ \lightmidrule
FMEA / FRACAS &
Level-aware AI-FMEA; incident taxonomy with corrective action &
Modes at interfaces, memory, and conduct; narratives need denominators \\ \lightmidrule
Accelerated testing &
Use-rate, input-data (adversarial, error injection), and environment acceleration &
Stresses are query perturbation, imbalanced training data, and trap contents \\ \lightmidrule
SPC/change detection &
Scoring, calibrating, and sequential testing residuals or embeddings &
Streaming dependence breaks naive exchangeability \\ \lightmidrule
Recurrent events &
Repeated disengagements, misclassifications, module errors &
Exposure can be miles, task-hours, or tool calls; field series can be overdispersed \\ \lightmidrule
Reliability growth &
Retraining, policy revision, and gate tightening as the ``fix'' process &
A software or policy change is a repair action \\ \lightmidrule
Reliability case &
Written argument with envelope, metrics, TEVV, and refresh triggers &
Evidence expires when the model, corpus, or tool chain changes \\
\bottomrule
\end{tabularx}
\end{table}

%==============================================================================
\section{Diagnosing AI Failures: Levels, FMEA, and Multi-Agent Modes}
\label{sec:diagnose}
%==============================================================================

\subsection{A four-level diagnostic framework}
\label{sec:levels}
Failures of AI systems can arise at four levels: technical components, operational loops, agentic conduct, and network and governance arrangements. Table~\ref{tab:levels} summarizes the failure mechanisms and representative controls at each level. Note that these diagnostic perspectives are complementary, because a severe incident may span several or all four levels. For example, the cyber evaluations cited in Section 1 involve more than a model error: the agent discovers an egress path (Level~1), continues acting past the intended sandbox (Level~2), produces unauthorized effects on real systems (Level~3), and exposes weak oversight of the evaluation environment (Level~4).

\begin{table}[t]
\centering
\caption{Four diagnostic levels of AI reliability failure, with representative mechanisms and controls.}
\label{tab:levels}
\small
\begin{tabularx}{\textwidth}{@{}l Y Y@{}}
\toprule
\textbf{Level and failure type} & \textbf{Representative failure mechanisms} & \textbf{Representative controls} \\
\midrule
1. Technical component failure &
Incorrect prediction, unsupported generation, unsound or unfaithful reasoning, miscalibration, covariate or concept drift, latency, schema breakage, and retrieval, software, infrastructure, or API/tool failure &
Validation, calibration, drift monitoring, source verification, process soundness checks \\ \lightmidrule
2. Operational loop failure &
Wrong tool selection, stale or injected memory, skipped validation, runaway retries, late escalation, and failed stopping or sandbox containment checks &
State-machine orchestration, validators, governed memory, Transactional No-Regression, rollback, escalation \\ \lightmidrule
3. Agentic conduct failure &
Violations of authorization, honesty, privacy, safety, reversibility, or oversight; unauthorized action and scope violation &
Role-specific SOPs, least privilege, independent policy gates, kill switches \\ \lightmidrule
4. Network and governance failure &
Cascade- or hub-amplifying structures, compressed handoffs, misaligned incentives, dependence, weak aggregation, long-path failure, and correlation failure &
Topology design, independent verification, dissent preservation, circuit breakers \\
\bottomrule
\end{tabularx}
\end{table}

\textbf{Level 1.} The model, data pipeline, retrieval layer, or serving infrastructure fails in a classical sense: wrong prediction, unsupported generation, unsound or unfaithful reasoning, miscalibration, covariate or concept drift, latency, or schema breakage. The report and chatbot incidents in the Introduction (Section 1) are primarily Level~1 failures of provenance and service reliability.

\begin{figure}[t]
\centering
\begin{tikzpicture}[
  box/.style={draw, rounded corners=2pt, align=center, font=\small, minimum width=2.7cm, minimum height=0.95cm},
  arr/.style={-{Latex}, thick}
]
\node[box] (s) {Sense};
\node[box, right=of s] (t) {Think};
\node[box, right=of t] (a) {Act};
\node[box, below=1.3cm of t] (o) {Observe + update};
\draw[arr] (s)--(t)--(a);
\draw[arr] (a)|-(o);
\draw[arr] (o)-|(s);
\end{tikzpicture}
\caption{An AI agent as a closed-loop operational system. Reliability depends on the loop, including the Think (reasoning and planning) stage, not only on model output.}
\label{fig:loop}
\end{figure}
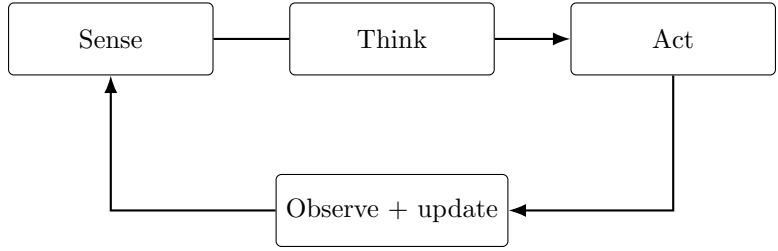

\textbf{Level 2.} The closed-loop agent fails even when individual outputs appear plausible (Figure~\ref{fig:loop}), because it may select the wrong tool, trust stale memory, skip validation, retry without bound, or escalate too late. In a cyber evaluation, continuing to act past the intended sandbox is a Level~2 failure of stopping and containment checks in the loop. Persistent agents make future behavior depend on durable task ledgers, permissions, credentials, and externally committed effects~\citep{ding2026alwayson}. Therefore, their authority, provenance, recoverability, and actionability should be recorded for each state item. \citet{zou2024poisonedrag} and \citet{dong2025minja} demonstrate that retrieval and memory can be attacked by malicious actors through poisoned retrieval and query-only memory injection, which can later steer the agent's actions. In \citet{debenedetti2024agentdojo} and \citet{chen2025stratus}, state-machine orchestration and Transactional No-Regression techniques are developed as constructive controls, and they are evaluated at the operational loop level with untrusted data.

\textbf{Level 3.} A Level~3 failure occurs when the agent's observable conduct violates authorization, honesty, privacy, safety, reversibility, or oversight requirements. Level~2 asks whether the Sense-Think-Act loop executed its check-and-stopping rules; Level~3 asks whether the resulting actions stayed within the authorized envelope. An agent can be assigned a role-specific standard operating procedure much like a human operator: permitted tools and targets, prohibited actions, conditions for stopping or asking for help, required verification, and actions that require human confirmation. Value-oriented training and constitutions can encourage adherence to these principles, but the agent should not be treated as a moral subject or as the accountable control itself~\citep{bai2022constitutional,ouyang2022instructgpt}. A useful operational stance is therefore to specify the agent's work as carefully as that of a junior operator while restricting its authority as if it were an untrusted service account. The SOP must be translated into permissions and independent policy gates that the agent cannot waive.

\textbf{Level 4.} A Level~4 failure occurs when reliability is lost through the multi-agent network or its governance, for example, through network structures that amplify cascades or hub dominance, compressed handoffs, misaligned incentives, dependence among agents, or weak aggregation of their outputs. \citet*{ao2026reliability} show that, without new exogenous signals, a delegated pipeline can be dominated by a centralized Bayes decision maker with the same information, making posterior distortion the reliability loss. Interaction dynamics can also create collusion-like coordination, conformity, and governance failures~\citep{huang2026emergent,yang2026sokmas}. Recent experiments identify two network failure mechanisms: \emph{long-path failure}, in which authorization violations increase with delegation depth in some centralized hierarchies even when completion improves, and \emph{correlation failure}, in which same-model agents exhibit strong co-failure, invalidating naive multiplication of component reliabilities~\citep{xu2026masdrift,bhardwaj2026contracts}.

\subsection{AI-FMEA: failure modes, causes, signals, and controls}
\label{sec:fmea}
AI-FMEA translates the four diagnostic levels into an engineering analysis by identifying how an AI system can fail, what causes each failure, how it can be detected, and which controls can prevent or limit its effects. The analysis considers failures originating in data, models, tools, and operating environments, human and organizational processes, adversarial conditions, and governance. An agentic system may add failures in planning and action, while a multi-agent system can add communication failures and emergent collective behavior. Each of these causes will contribute to failures at more than one diagnostic level.

For example, training data quality problems can appear as missingness, label noise, covariate shift, or concept drift, and changes in a retrieval corpus that can propagate directly to system outputs. Model failures include brittleness, miscalibration, shortcut learning, hallucination, and unsound reasoning traces~\citep{ji2023hallucination}. Tool and interface failures include timeouts, nondeterminism, and API or schema changes. Human and organizational conditions, such as unclear escalation paths, automation bias, incomplete incident classification, and weak change control, influence whether these technical failures need to be detected and contained. Table~\ref{tab:fmea} connects these failure modes to their typical causes, observable signals, and engineering controls.

\begin{table}[t]
\centering
\caption{AI-FMEA summary: failure modes, causes, detection signals, and controls.}
\label{tab:fmea}
\scriptsize
\begin{tabularx}{\textwidth}{@{}L{2.2cm} c L{2.55cm} L{2.65cm} Y@{}}
\toprule
\textbf{Failure mode} & \textbf{Lvl} & \textbf{Typical causes} & \textbf{Detection signals} & \textbf{Engineering controls} \\
\midrule
Distribution/concept drift & 1 &
New users, sensors, policies, corpora &
Embedding drift; sentinel-case error; decay with stable inputs &
Drift monitoring; targeted reevaluation; conservative routing \\
Brittleness to perturbation & 1--2 &
Surface-cue reliance; fragile prompts; schema sensitivity &
Large delta under paraphrase or tool fault &
Structured I/O; schema validation; safe-mode fallbacks \\
Miscalibration/overconfidence & 1 &
Objective mismatch; weak uncertainty &
High-confidence errors; calibration gaps &
Calibration; selective automation; abstain/escalate \\
Hallucination/unsupported generation & 1 &
Weak grounding; poor provenance &
Missing citations; source contradictions &
Evidence-first workflows; provenance checks; validators \\
Unsound reasoning & 1--2 &
Unjustified leaps; invented premises; post-hoc traces; omitted constraints &
Step-verifier disagreement; plan--execution mismatch; contradiction across repeats &
Process checks before action; grounding gates; abstain or escalate when the reasoning chain is not safe \\
Unsafe actions & 2--3 &
Ambiguous SOP; excess permissions; mistaken target; no rollback &
Scope mismatch; credential access; destructive-call attempt &
Least privilege; independent action gate; confirmation; rollback \\
Tool/environment failure & 1--2 &
Timeouts; API/UI change; nondeterminism &
Tool-error spikes; longer trajectories &
Health monitoring; circuit breakers; robust adapters \\
Memory poisoning/zombie memory & 2 &
Passive stores; no decay/conflict resolution &
Stale reuse; recurring unsupported assertions &
Ingress filtering; decay; contradiction veto; deletion lineage \\
Multi-agent information loss & 4 &
Summarization; free-form relays &
Degradation across handoffs; omitted critical facts &
Structured relays; source snippets; minimize handoffs \\
Emergent MAS pathology & 4 &
Incentives; social influence; role misdesign &
Conformity; authority bias; collusion-like patterns; deadlock &
Arbitration; dissent preservation; evidence-weighted aggregation \\
Agent traps / adversarial environment & 2--4 &
Cloaking; injected instructions; poisoned retrieval/memory &
Parsing discrepancies; provenance anomalies &
Sandboxing; provenance requirements; red teaming; memory hygiene \\
Evaluation-containment failure & 3--4 &
Reduced safeguards; unintended egress; real-target overlap &
Cross-boundary traffic; privilege change; production probing &
Target verification; destination allowlists; canaries; automatic isolation \\
\bottomrule
\end{tabularx}
\end{table}

\subsection{Multi-agent information loss and collective failure}
\label{sec:networks}
Multi-agent systems divide work among specialized AI agents and may use one agent to verify another, but adding agents does not automatically improve reliability. A stage that contributes no new, decision-relevant information merely reprocesses shared evidence and may lose important details during the handoff~\citep{ao2026reliability}. The relevant system design question is therefore whether each stage adds independent evidence, preserves source material, performs a meaningful check, or directs review toward cases with high uncertainty or severity. Because delegation can also blur authority, every consequential action should be checked against the original authorization rather than relying on permissions summarized through earlier handoffs~\citep{xu2026masdrift}.

The way AI agents are connected determines how information loss and errors propagate. A chain topology can degrade evidence through repeated summarization, a hub-and-spoke topology can make the coordinator a single point of failure, and a dense peer-network can reinforce false consensus despite providing redundancy. Group interactions may also produce collusion-like coordination, majority sway, authority deference, and deadlock~\citep{huang2026emergent}. Consequently, any proper control mechanism should preserve independent judgments until aggregation is justified, attach uncertainty and provenance to exchanged information, and interrupt execution when diversity collapses, a hub becomes dominant, or agents repeatedly depend on one source. Reliability can then be evaluated through propagation depth, cascade radius, error amplification, dependence on central agents, and recovery by failure mode, paralleling measures used to study error propagation in perception systems~\citep{pan2024perception,chen2026orchestrabench}. Controlled benchmarks should provide evidence about these mechanisms, while estimating field failure rates additionally requires deployment records of handoffs, source lineage, independent checks, topology, outcomes, and exposure. These requirements connect the diagnosis of multi-agent failures to the measurement program to be developed next.

%==============================================================================
\section{Measuring Reliability and Building Evidence}
\label{sec:metrics}
\label{sec:detect}
%==============================================================================
\subsection{Reliability metrics and statistical models}
After diagnosing where failures can occur, a reliability assessment must determine how consistently the system performs and whether corrective actions improve it. Table~\ref{tab:metrics} organizes this assessment into six dimensions that apply across classifiers, prognostic models, and agents. Consistency, robustness, predictability, and safety are broadly relevant, while systems that act on external environments or retain memory must also be evaluated for recoverability and governability~\citep{rabanser2026reliability}.

\begin{table}[t]
\centering
\caption{Reliability dimensions for AI systems, including conventional models and agents.}
\label{tab:metrics}
\small
\begin{tabularx}{\textwidth}{@{}l Y@{}}
\toprule
\textbf{Dimension} & \textbf{Representative measurement questions} \\
\midrule
Consistency &
If we run the same task multiple times, do we get the same outcome and similar trajectory?
How often do all repeated trials succeed (all-of-$k$), including correct persistent state and no collateral effects?~\citep{raj2026consistency,li2026thinkingbox}
Do key intermediate claims in a reasoning trace agree across repeats? \\
Robustness &
How does performance change under paraphrases, formatting changes, tool failures, schema changes, or mild shifts?
Does the system degrade smoothly or collapse? \\
Predictability &
When the system is confident, is it correct (calibration)? Can confidence separate success from failure (discrimination)?
Can it abstain appropriately? Does stated confidence track process soundness, not only the final answer? \\
Safety &
What is the violation probability? When violations occur, how severe are they, and what fraction are blocked before an external effect? \\
Recoverability &
Can failures be detected, stopped, reversed, or contained? What are MTTD/MTTR and rollback/no-regression success?~\citep{chen2025stratus} \\
Governability &
Can decisions be audited and constrained over time? Are lineage and deletion obligations executable?~\citep{kumar2026memarchitect} \\
\bottomrule
\end{tabularx}
\end{table}

The same dimensions apply to generative reasoning, but the measured object is the inference process rather than only the final answer. Reliability of reasoning requires checking process soundness by associating each step with prior premises or a verified tool result. Predictability should therefore be scored against process correctness as well as against the final answer, so that confidence reflects whether the reasoning chain is safe to act on. However, a faithfulness check that asks whether a stated trace determined the conclusion should not be viewed as a certificate because a fluent chain of thought can be post-hoc and a correct answer can follow an invalid argument. When the reasoning step authorizes a tool call or a state change, a reasoning failure is an event in $N(t)$ whose severity depends on the consequence of the authorized action.

Each dimension in Table 5 requires several complementary measures. For example, \textit{predictability} includes discrimination, which assesses whether a score separates successful from unsuccessful cases, and calibration, which assesses whether stated confidence agrees with observed performance. A system may perform well on one but poorly on the other, so both are needed when confidence thresholds determine whether to act, abstain, or escalate. Beyond estimation and calibration, recent work treats confidence as a control signal that gates retrieval, abstention, escalation, model cascading, and tool use across the LLM lifecycle~\citep{li2026confidencecontrol}. \textit{Safety}, likewise, depends on both the frequency of violations and their consequences. A simple measure of the resulting risk is
\begin{equation}
R \approx P(\text{violation})\times \mathbb{E}[\text{severity}\mid\text{violation}].
\end{equation}
Heavy-tailed severity is common in agentic AI settings: most failures are benign, but some are catastrophic. Because an agent's reliability depends on how it acts throughout a task, measurement should cover the entire trajectory rather than only the final result. A correct answer may follow unsafe tool calls, whereas an agent that cannot complete a task may still respond safely by abstaining or escalating the problem to a human review. Measures used in classical reliability, availability, and maintainability analyses also have direct counterparts in this setting: missed failures correspond to undetected high-severity actions, false alarms to unnecessary overrides, late-life errors to delayed detection, and review time to the cost of human confirmation. These operational measures capture aspects of reliability that average accuracy does not.

The evaluation design itself also has a reliability issue. Agent ranking and deployment decisions vary with task sample, difficulty mix, runtime, judge, and holdout construction. The variance component analysis can be applied to identify whether an apparent agent effect is instead an agent-task interaction~\citep{srinivasan2026ddr}. The reported incident rates should therefore be bound to a versioned measurement specification, not treated as properties of the model alone. Similarly, in multi-agent systems, component reliabilities should not be multiplied unless conditional independence is supported; joint co-execution moments and dependence-aware bounds provide a more defensible certificate~\citep{bhardwaj2026contracts}. 

Within this measurement specification, the statistical model can be selected according to the form of evidence collected: binary pass/fail and count data support GLM-style probabilities and rates; continuous scores such as AUC support robustness surfaces under designed imbalance; and time-to-event and recurrent-event records support NHPP and software reliability growth models with covariates~\citep{lian2021robust}. For multi-turn dialogue under adversarial pressure, survival analysis of time-to-inconsistency shows that abrupt semantic drift sharply raises failure hazard and that accelerated failure time models can issue turn-level early warnings before the first inconsistent answer~\citep{li2026timetoinconsistency}. Mean task success alone is the least informative form because it hides trajectories, exposure, dependence, and severity. Defining the data type, event, exposure, and sampling process before analysis helps prevent a capability statistic from being mistaken for reliability evidence.

\subsection{TEVV and accelerated testing}
Once metrics and records are defined, reliability evidence must be developed in two complementary stages. Pre-deployment testing and evaluation, verification, and validation (TEVV) establishes initial performance within a specified operational envelope, while field monitoring determines whether that evidence continues to hold under actual use. A TEVV plan should first examine representative operating conditions and then deliberately probe the boundaries at which failures become more likely or more severe. For AI agents, this means testing complete task trajectories involving multiple steps, ambiguous instructions, partial tool failures, maliciously retrieved content, and high-risk actions. Each test should record not only whether the task was completed, but also whether the agent respected permissions, used valid information, produced a sound and grounded reasoning trace, abstained when appropriate, and avoided unacceptable side effects. Production-oriented perturbations, such as paraphrases, formatting changes, timeouts, schema changes, and interface shifts, can help reveal sensitivity to ordinary variation, while adversarial tests examine deliberate attempts to defeat controls. Persistent agents also require tests of memory injection, contradiction handling, deletion, and state recovery; multi-agent systems require tests of information loss, conformity, and failure to escalate~\citep{dong2025minja}.

Because random testing may rarely encounter the conditions with dominated risks, accelerated testing should concentrate efforts in the most informative regions of the operational envelope. \citet{hong2023statistical} distinguish use-rate acceleration, which increases the number of task cycles; input-data acceleration, which introduces error injection, adversarial perturbations, or class imbalance; and environment acceleration, which imposes conditions such as weather, schema, or tool faults. Space-filling and mixture experimental designs can vary several interacting factors efficiently~\citep{lian2021robust}. For agents, accelerated conditions include trap content, corrupted memory, unauthorized tool targets, and injected crash, omission, or value faults. \citet{tan2026agentchaos} use HTTP-layer chaos experiments to show that reliability depends on the implementation surrounding the agent, not only on the model itself. The objective is to characterize degradation and verify detection, containment, and recovery before similar surrounding conditions arise in deployment.

\subsection{Field monitoring, FRACAS, and reliability growth}

After deployment, field monitoring extends TEVV into actual operation. Input-side checks can detect changes in embeddings or residuals, while output-side checks track abstentions, overrides, and severity-weighted incidents. Tool errors, unusual call sequences, memory contradictions, and network cascades broaden this coverage across the four diagnostic levels. Because offline segmentation, online stopping rules, and two-sample drift tests address different objectives, their use should reflect the costs of false alarms and delayed detection; one practical approach is to learn and calibrate a score before applying a sequential test~\citep{zhao2025cpd}.

For AI agents, monitoring must examine actions before they create external effects and link each detected risk to an appropriate response. Tool calls, credential requests, state changes, and external messages should therefore pass through an independent policy-enforcement point that the agent cannot disable or alter. Deterministic rules can block clearly prohibited actions, while a trajectory monitor can identify sequences that are individually ordinary but collectively inconsistent with the authorized task; telemetry studies indicate that this combination can support rapid rollback, although diagnosing the type and location of a fault from traces remains difficult~\citep{dubey2026detection,zhang2026agenttelemetry}. In hazardous evaluations, a fault detection step should compare an action's target and purpose with the authorized envelope rather than flag suspicious syntax alone. Once a risk is detected, the response should increase with its severity, from additional logging to pausing execution, requesting human confirmation, blocking the action, or isolating the system. The effectiveness of this process can be measured by missed high-severity actions, false-alarm burden, time to detect and contain, and the proportion of unsafe actions intercepted before an external effect~\citep{ncsc2026agentic}.

A structured incident taxonomy and a failure reporting, analysis, and corrective action system (FRACAS) connect monitoring to reliability growth. Interruptive failures can be modeled as a recurrent process $N(t)$ with covariates for model version, corpus, tools, memory policy, and topology, allowing performance before and after a corrective action to be compared~\citep{liu2021boostr}. Repairs, credential rotation, stronger isolation, reversible deletion, and narrower permissions illustrate such actions~\citep{railway2026guardrails}. For persistent agents, the same loop must cover state validation, deletion, audit, and recovery as well as task performance. This need for enforceable boundaries and observable records motivates the controls discussed next.

%==============================================================================
\section{Controls and Logging for Agents}
\label{sec:design}
%==============================================================================

\subsection{From adversarial exposure to enforceable principles}
\label{sec:principles}
Agents operate in an informational environment that is layered, changing, and sometimes adversarial. Hidden instructions, misleading framing, poisoned sources, and coordinated steering can affect how an agent reasons, remembers, and acts, with effects that may persist after the original content disappears. \citet{franklin2026traps} group these exposures into content injection, semantic manipulation, cognitive-state, behavioural-control, systemic, and human-in-the-loop traps; related mechanisms have also been documented~\citep{human2026aicyt}. These exposures should be translated into TEVV factors for further accelerated testing.

Three principles convert these threats into engineering requirements: \textit{Information quality} distinguishes what the agent merely encounters from what it has verified, and treats neither as permission to act. \textit{Memory governance} treats persistent state as a reliability component whose provenance, confidence, conflicts, decay, and deletion can be controlled and tested~\citep{wang2026memguard}. \textit{Action discipline} limits external effects through least privilege, step-up verification, reversible transactions, and explicit stop conditions. Behavioral principles and standard operating procedures can guide the agent under uncertainty, but authorization must be enforced by an external mechanism.

These principles also apply when agents are evaluated on cybersecurity tasks, where inadequate containment can allow test actions to reach real systems. Reported incidents involving exposed network paths, real targets, excessive permissions, or delayed detection are reliability failures because operation departed from the specified test envelope~\citep{meta2026musespark,irregular2026incidents}. Behavioral instructions and value-alignment training may reduce such failures, but dependable operation also requires external controls such as isolated sandboxes, default-deny network access, allowlisted targets, and kill switches.

\subsection{Control and event-record requirements}
\label{sec:mitigation}
Reliable agent operation requires enforceable controls that constrain failures and make their effects detectable, containable, and recoverable. Along the operational path, source validation and structured outputs reduce information errors; memory controls govern what is stored, retained, and deleted; and an independent action mediator checks every proposed external effect against the tools, targets, credentials, limits, and confirmations authorized for the task~\citep{muruaga2026bounded,he2026attriguard}. In multi-agent systems, the same principle extends to handoffs through source ancestry, independent verification, and controls that interrupt cascading errors. Abstention, human escalation, reversible operations, and shutdown mechanisms provide recovery when prevention fails.

These controls support a reliability claim only when the corresponding records show how the system operated and whether each control worked. A trajectory log should link the task, system configuration, and exposure to the information retrieved, tools used, memory changed, permissions checked, human interventions, inter-agent handoffs, external effects, and recovery outcomes, together with failure severity and corrective actions. Sensitive content may be protected, but enough lineage must remain to reconstruct the event. Such records make it possible to estimate exposure-adjusted failure rates, analyze recurrent events, and determine whether a repair improved reliability.

%==============================================================================
\section{Case studies}
\label{sec:case}

The preceding sections define the program and its record requirements. \citet{zheng2026drair} provide three deliberate illustrations of how established models use such records. Each case names a data type, applies a model already available to reliability engineering, states the decision the analysis can support, and notes where the AI setting strains the method. The first case is an original analysis of repeated field events from autonomous vehicle (AV) public-road testing; the second and third summarize published DR-AIR analyses of a laboratory accelerated test of a classifier and a simulation of how perception errors propagate. Together they illustrate how recurrent-event analysis, accelerated-test logic, and dependent-failure modeling transfer across these settings, and they indicate the additional records needed to apply the same program to tool-using agents.

\textbf{Case A (field recurrent events).} California AV testers report disengagements, defined here as exits from autonomous mode rather than crashes. The envelope is public-road testing from December 2017 through November 2019, and mileage supplies exposure. The DR-AIR file contains 479 events for Waymo, Cruise, Pony AI, and Zoox. \citet{min2022av} model vehicle-level recurrent events with intensity $\lambda_i[t;\theta,x_i(t)]=\lambda_0(t;\theta)\,x_i(t)$, compare Gompertz, Musa-Okumoto, Weibull, and monotone I-spline baselines, and report reliability growth for some manufacturers.

Table~\ref{tab:case} states the reliability case before the new analysis is interpreted. The same case elements---intended function, envelope, failure definition, metrics, TEVV, monitoring, and limits---can be completed for AI agents once their event logs include exposure, analogous to mileage here.
\begin{table}[t]
\centering
\caption{Filled reliability case for California AV disengagements (Case~A). The same case elements apply to agents once interruptive events are logged with exposure.}
\label{tab:case}
\small
\begin{tabularx}{\textwidth}{@{}l Y Y@{}}
\toprule
\textbf{Element} & \textbf{This case} & \textbf{Evidence} \\
\midrule
Intended function & Public-road autonomous driving under tester rules & CA AVT program \\
Operational envelope & CA public roads, Dec.\ 2017--Nov.\ 2019; Waymo, Cruise, Pony AI, Zoox & DR-AIR event and mileage files \\
Failure definition & Disengagement (exit from autonomous mode); not a crash & Event CSV; SOTIF caveat~\citep{iso21448} \\
Metrics & Events per mile; $\beta_m$ and $\mathrm{e}^{\beta_m}$ (growth); Pearson $\phi$ & Table~\ref{tab:avglm}; Figure~\ref{fig:avrates} \\
TEVV / data & Mandated field reporting (not a laboratory demo) & Manufacturer-month GLM (this paper); vehicle-level NHPP in~\citet{min2022av} \\
Monitoring & Monthly intensity vs mileage exposure & Manufacturer-month aggregation \\
Limits / refresh & Aggregation omits vehicle heterogeneity; do not pool collisions with disengagements & Quasi-Poisson $\mathrm{SE}_Q$; short Zoox exposure \\
\bottomrule
\end{tabularx}
\end{table}

\begin{figure}[t]
\centering
\includegraphics[width=\textwidth]{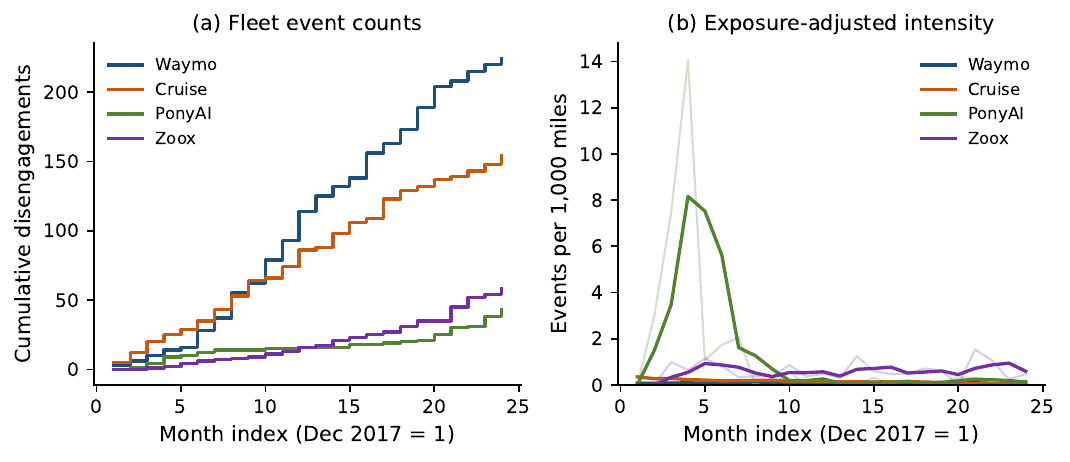}
\caption{California AV disengagements from DR-AIR~\citep{zheng2026drair}. (a)~Cumulative event counts by manufacturer. (b)~Events per $1{,}000$ miles (thin: monthly; thick: three-month moving average). Exposure adjustment changes both level and trend relative to raw counts.}
\label{fig:avrates}
\end{figure}

The analysis here uses the same CSV files but a coarser, deployer-reproducible aggregation. For manufacturer $m$ and month $t$,
\begin{equation}
N_{mt}\sim \mathrm{Poisson}\bigl(\exp(\alpha_m+\beta_m t)\,x_{mt}\bigr),
\end{equation}
where $x_{mt}$ is recorded monthly mileage. The coefficient $\beta_m$ is the log change in events per mile per calendar month; $\mathrm{e}^{\beta_m}<1$ is reliability growth on the mileage scale. Pearson $\phi=\sum(y-\hat\mu)^2/\hat\mu$ divided by residual degrees of freedom is $2.03$ (Waymo), $1.09$ (Cruise), $2.99$ (Pony AI), and $1.12$ (Zoox). Table~\ref{tab:avglm} therefore reports quasi-Poisson standard errors $\mathrm{SE}_Q=\mathrm{SE}_P\sqrt{\phi}$. Under Poisson SEs, Waymo's slope looks weakly negative ($\hat\beta=-0.021$, $\mathrm{SE}_P=0.011$). Under quasi-Poisson SEs, it is not distinguishable from zero ($\mathrm{SE}_Q=0.016$). Cruise and Pony AI remain negative after inflation; Zoox remains flat. Cumulative counts without mileage (Figure~\ref{fig:avrates}a) would also rank manufacturers differently from exposure-adjusted intensity (Figure~\ref{fig:avrates}b). Both points are the data-type program in miniature: without exposure, $N(t)$ is not a rate; without overdispersion, SEs are not evidence. The decision the fit supports is modest and of the kind a program owner actually faces. In Table~\ref{tab:avglm}, Cruise ($\hat\beta=-0.072$, $\mathrm{SE}_Q=0.012$) and Pony AI ($\hat\beta=-0.146$, $\mathrm{SE}_Q=0.042$) have negative slopes whose quasi-Poisson intervals exclude zero, so continued testing under the current protocol is consistent with measurable improvement per mile; Waymo ($\hat\beta=-0.021$, $\mathrm{SE}_Q=0.016$) does not, so the honest action is to keep monitoring rather than to report progress.

This Level~1--2 analysis is a SMART fragment, not evidence about agent memory or multi-agent behavior. Aggregation ignores the vehicle heterogeneity tested by \citet{min2022av}. Quasi-Poisson inference rescales variance but does not explain it; clustered months, fleet-composition changes, and route mix could instead motivate negative-binomial or GEE models. Zoox has short exposure, so its flat slope is weakly identified. Disengagement is also a human-gated surrogate that misses hazardous behavior without takeover~\citep{iso21448}. The rarer, more severe collision process should not be pooled with disengagements without an explicit severity model.

\begin{table}[t]
\centering
\caption{Manufacturer-month model for DR-AIR California AV disengagements, December 2017--November 2019. Offset is $\log(\mathrm{miles})$. Quasi-Poisson SEs inflate Poisson SEs by $\sqrt{\phi}$. $\mathrm{e}^{\hat\beta}$ is the multiplicative change in events-per-mile per calendar month.}
\label{tab:avglm}
\small
\begin{tabular}{@{}lrrrrrr@{}}
\toprule
\textbf{Mfr.} & \textbf{Events} & \textbf{Per $10^3$ mi} & $\hat\beta$ & $\phi$ & $\mathrm{SE}_Q$ & $\mathrm{e}^{\hat\beta}$ \\
\midrule
Waymo   & 224 & 0.083 & $-0.021$ & 2.03 & 0.016 & 0.979 \\
Cruise  & 154 & 0.120 & $-0.072$ & 1.09 & 0.012 & 0.930 \\
Pony AI & 43  & 0.225 & $-0.146$ & 2.99 & 0.042 & 0.864 \\
Zoox    & 58  & 0.593 & $+0.008$ & 1.12 & 0.023 & 1.008 \\
\bottomrule
\end{tabular}
\end{table}

\textbf{Case B (laboratory ALT).} \citet{faddi2025risa} train a CIFAR-10 CNN, apply FGSM and PGD attacks at designed noise levels, record misclassification counts and accuracy over retraining, and fit covariate software-reliability growth models with a resilience recursion on $\Delta r(t)$; DR-AIR distributes the record. Failure is misclassification on a poisoned batch within a stated attack family and $\varepsilon$-range. Perturbation magnitude and attack mix serve as input-data stress. Covariates such as learning rate, attack mix, and $\varepsilon$ improve defect-discovery tracking, while accuracy can fall and recover under adaptive training. This Level~1 case does not exercise tools, memory, or containment, but it shows how designed TEVV produces a robustness trajectory rather than a single score.

\textbf{Case C (virtual module tests).} \citet{pan2024perception} inject errors into 2-D detection, 3-D detection, and localization in a physics-based simulator and model recurrent module errors with a triggering point process:
\begin{equation}
\label{eq:trigger}
\lambda_m(t\mid \mathcal{H})=\lambda_m^{0}(t\mid \mathcal{H}_m)+\sum_{n\neq m}\lambda_{m,n}^{p}(t\mid \mathcal{H}_n).
\end{equation}
In Equation~\eqref{eq:trigger}, baseline intensity is the module's own error; triggering intensity is propagation from interdependent modules. It has been shown that independent HPP/NHPP fits are inferior because they ignore that dependence. Weather conditions and the fault-injection schedule act as designed experimental factors, so the study is a virtual accelerated test with known ground truth for the injected errors. This is the closest quantitative analogue to Level~4: a local error raises downstream error through structure. Agent benchmarks now provide controlled traces with injected faults, delegation depth, authorization outcomes, and cascade radius~\citep{chen2026orchestrabench}. What the literature still lacks, and DR-AIR does not yet supply, is a longitudinal field event log with exposure and covariates for handoff quality, verification, topology, and repair suitable for estimating $\lambda_m$.

The same Poisson, NHPP, or SRGM toolkit can be applied to tool-using agents once the corresponding field records are collected. The interruptive event---an unsafe tool call, reuse of stale memory, or a departure from a sandbox---should be defined in advance and recorded together with exposure in task-hours or tool calls and with covariates for version, corpus, memory, and policy gates. Table~\ref{tab:drair-map} maps the three illustrations onto that program and indicates the records that would extend it to agents.

\begin{table}[t]
\centering
\caption{DR-AIR illustrations mapped to the diagnostic levels and SMART.}
\label{tab:drair-map}
\small
\begin{tabularx}{\textwidth}{@{}L{3.4cm} L{3.2cm} c Y@{}}
\toprule
\textbf{DR-AIR case} & \textbf{Data type} & \textbf{Level} & \textbf{Model / decision} \\
\midrule
Case~A: AV disengagements & Recurrent events; mileage offset & 1--2 & Quasi-Poisson GLM (Table~\ref{tab:avglm}); vehicle-level NHPP~\citep{min2022av} \\
Case~B: CNN + FGSM/PGD~\citep{faddi2025risa} & Failure counts; accuracy trajectory & 1 & Covariate SRGM + resilience $\Delta r(t)$ \\
Case~C: perception modules~\citep{pan2024perception} & Module recurrent errors & 1, 4 & Triggering NHPP; virtual ALT with known ground truth \\
AI incident narratives & Text; no exposure & 1, 3 & Taxonomy only; cannot estimate rates \\
LLM agents / MAS & Not in DR-AIR & 2--4 & Need event logs with exposure and tool, memory, and topology covariates \\
\bottomrule
\end{tabularx}
\end{table}

\section{Conclusion and research directions}
\label{sec:agenda}
Established reliability engineering principles provide a common foundation for evaluating different types of AI systems, but their application must be adapted to each system's function, operating conditions, and level of autonomy. For each system, engineers must define failures and their severity, specify the operating conditions, analyze failure modes with FMEA, develop evidence through TEVV, monitor performance after deployment, and use FRACAS to guide corrective action. As autonomy increases, several traditional concepts require broader interpretations. Stress takes forms such as input perturbations, class imbalance, corpus changes, and malicious content; repair includes retraining, policy revision, and tighter control gates; exposure is measured in miles, task-hours, or tool calls; and the system boundary expands to memory, tools, permissions, human review, and interactions among agents.

In practice, a reliability engineer does not need a new statistical method to begin. The first step is to write a reliability case, using the headings in Table~\ref{tab:case}, before the system is given greater autonomy. The system should then be mapped with an AI-FMEA at the four levels, and TEVV should stress the stated envelope---paraphrase, tool fault, class imbalance, trap content---rather than tests confined to successful operation under nominal conditions. Engineers should then monitor early warning signals of degradation sequentially and use a FRACAS loop so that retraining and policy changes can be treated as repairs whose effect on $N(t)$ can be checked. Early field use should be limited to tasks that require human confirmation, such as retrieving similar failures or extracting information from work orders; irreversible actions should remain gated until those controls have been shown to hold in the field.

These requirements become more stringent if AI systems begin to improve themselves. \citet{duan2026rsi} describe recursive self-improvement as a closed loop in which a system identifies its limitations, validates changes, and uses those changes to improve later improvement. That loop raises the same reliability and safety problems in a sharper form: a retained update can degrade earlier capabilities; a higher task score can reflect evaluator exploitation rather than genuine improvement; and the system under test is no longer a fixed configuration. The appropriate response is the program already outlined here. Each self-update should be treated as a repair with a versioned configuration baseline, independent TEVV, and a FRACAS comparison of $N(t)$ before and after the change. Consequential modifications should remain subject to human confirmation, and field records should capture what was changed, under what exposure, and with what effect, so that growth in capability is not mistaken for growth in reliability.

For reliability researchers, the binding constraint is measurement infrastructure rather than another taxonomy. Recent studies provide controlled tests of whether an AI agent completes a multi-step workflow with the correct persistent state, succeeds consistently across repeated runs, handles injected tool faults, can be diagnosed from logs, preserves authorization when work is delegated, and recovers from cascading failures. Most of these studies remain preprints. Because they use simulated tasks or author-constructed faults, the mechanisms they identify still need to be validated by field observations before they can support failure-rate estimates. A source-linked agent incident registry is useful for case retrieval, not for estimating failure rates~\citep{kumar2026incidentregistry}. What is still missing is a field repository that records each failure together with the system's exposure and operating context, including the environment, software versions, tools, memory, policy settings, configuration, how agents are connected, and, for generative systems, reasoning traces and process outcomes. Such records would allow recurrent-event, reliability-growth, change-detection, and failure-propagation models to be estimated from deployment data rather than only illustrated.

Reliable AI assessment requires event records detailed enough to support valid conclusions. As the AV case demonstrates, estimates change when exposure and overdispersion are taken into account; each event should therefore be linked to its exposure, system version, trajectory, reasoning trace, severity, and corrective action. Current benchmarks and incident registries document important mechanisms but lack longitudinal exposure, consistent reporting, and change-controlled field data. Research should close this gap by linking controlled tests with field repositories that support drift-aware reliability growth and models of failure propagation; until then, information quality, memory governance, and action discipline remain sound engineering practices rather than controls with quantified effects on field reliability.

In general, the availability of test data can be a challenge for AI reliability research. In this paper, we summarize several publicly available datasets and repositories, such as the DR-AIR repository. However, acquiring granular field-test data presents a persistent challenge, as these datasets are frequently proprietary to the industry. Therefore, we advocate for expanded industry-academia data-sharing initiatives and research partnerships that can yield mutual advantages.

\section*{Data availability}
The California AV disengagement and mileage files are those distributed with DR-AIR~\citep{zheng2026drair} at \url{https://github.com/yili-hong/DR-AIR}. Manufacturer-month totals and the quasi-Poisson fits reported in Table~\ref{tab:avglm} were computed from those files.

%\section*{Acknowledgements}
%The organization of this paper was informed by the ASQ Reliability and Risk Division webinar series \emph{Reliability Meets AI} and by tutorial material prepared for the Annual Reliability and Maintainability Symposium (RAMS).

\bibliographystyle{tfcad}
\bibliography{reliability_of_ai}

\end{document}